\documentclass{article}

\usepackage{arxiv}

\usepackage[utf8]{inputenc} 
\usepackage[T1]{fontenc}    
\usepackage{hyperref}       
\usepackage{url}            
\usepackage{booktabs}       
\usepackage{amsfonts}       
\usepackage{nicefrac}       
\usepackage{microtype}      
\usepackage{lipsum}

\usepackage{bm}
\usepackage{amsmath}
\usepackage{amssymb}
\usepackage{amsfonts}
\usepackage{graphicx}
\def\vec#1{\mathbf{#1}}
\usepackage{algorithm}
\usepackage{algorithmic}
\usepackage{comment}

\title{Few-shot Learning for Topic Modeling}

\author{
  Tomoharu Iwata\\
  NTT Communication Science Laboratories\\
}
\date{}

\begin{document}
\maketitle

\begin{abstract}
Topic models have been successfully used for analyzing text documents. However, with existing topic models, many documents are required for training. In this paper, we propose a neural network-based few-shot learning method that can learn a topic model from just a few documents. The neural networks in our model take a small number of documents as inputs, and output topic model priors. The proposed method trains the neural networks such that the expected test likelihood is improved when topic model parameters are estimated by maximizing the posterior probability using the priors based on the EM algorithm. Since each step in the EM algorithm is differentiable, the proposed method can backpropagate the loss through the EM algorithm to train the neural networks. The expected test likelihood is maximized by a stochastic gradient descent method using a set of multiple text corpora with an episodic training framework. In our experiments, we demonstrate that the proposed method achieves better perplexity than existing methods using three real-world text document sets.
\end{abstract}

\section{Introduction}

There is great interest in topic modeling, such as latent Dirichlet allocation~\cite{blei2003latent}
and probabilistic latent semantic analysis~\cite{hofmann1999probabilistic}, for analyzing text documents,
which include scientific papers, news articles, and web pages~\cite{blei2012probabilistic,blei2006dynamic,griffiths2004finding,mcauliffe2008supervised,iwata2009topic,mimno2011bayesian,boyd2017applications,doyle2009accounting,zhao2011comparing,chang2009reading,mcauliffe2008supervised}.
A topic model is a probabilistic generative model, in which a document is modeled as a mixture of topics,
and a topic is modeled as a probability distribution over vocabulary terms.
Topic models have been successfully used in a wide variety of applications including information retrieval~\cite{blei2003latent},
collaborative filtering~\cite{hofmann2003collaborative},
visualization~\cite{iwata2008probabilistic},
and entity disambiguation~\cite{kataria2011entity}.

For estimating topic models, many documents are required.
However, some applications may have insufficient numbers of documents.
In this paper, we propose a few-shot learning method for topic modeling,
which can estimate topic models from just a small number of documents.
Although many few-shot learning methods have been proposed~\cite{schmidhuber:1987:srl,bengio1991learning,ravi2016optimization,andrychowicz2016learning,vinyals2016matching,snell2017prototypical,bartunov2018few,finn2017model},
existing few-shot learning methods are not intended for topic modeling.

\begin{figure}[t]
  \centering
  \includegraphics[width=22em]{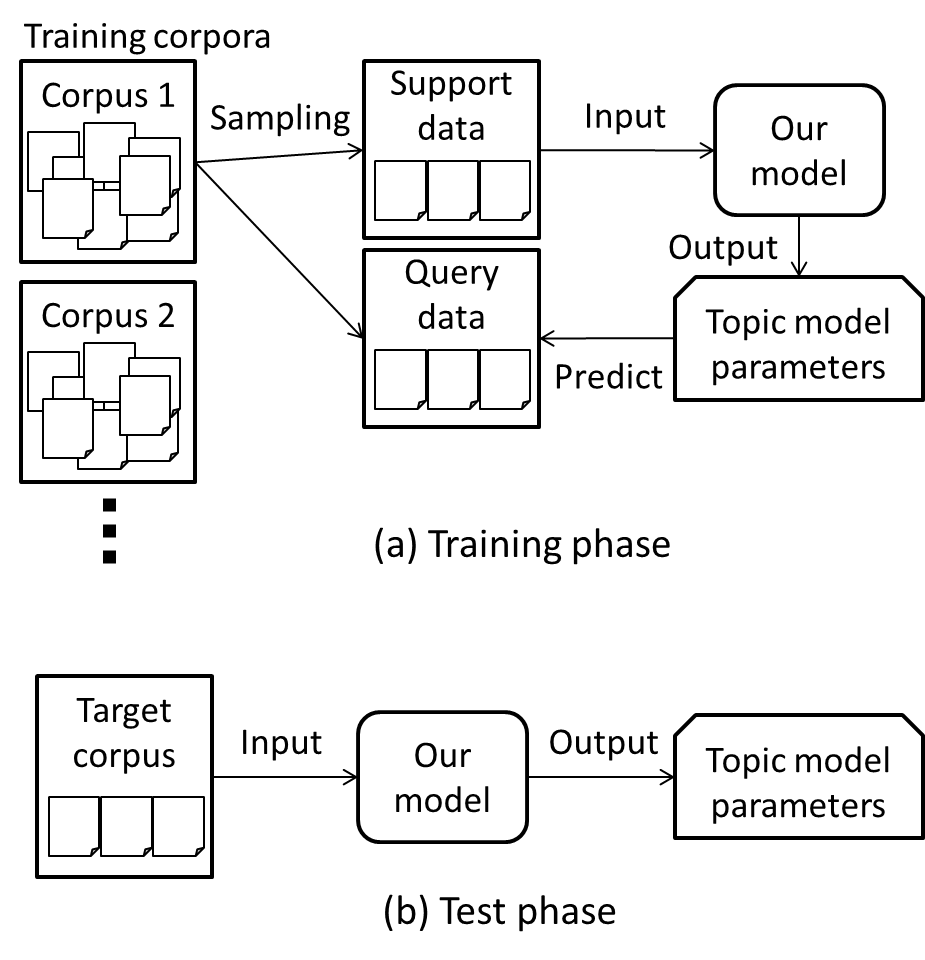}
  \caption{Our framework. (a) In the training phase, our model is trained using multiple corpora. For each corpus, support and query data are sampled. Our model takes the support data as input, and outputs topic model parameters. The topic model parameters are used to predict the query data, by which we calculate the loss to train our model. (b) In the test phase, our model generates topic model parameters given a small number of documents in the target corpus.}
  \label{fig:method}
\end{figure}

The proposed method uses a neural network-based model that generates topic model parameters.
In the training phase, we are given a set of multiple text corpora,
where each corpus is assumed to be generated from a topic model with corpus-specific parameters.
We train our model by maximizing the expected test log-likelihood using the training corpora with an episodic training framework~\cite{santoro2016meta},
where support and query data are randomly sampled
from training corpora to simulate the test phase for each epoch.
The support data are used for generating a corpus-specific topic model,
and the query data are used for evaluating the generated topic model.
In the test phase, we are given a small number of documents,
which are support data, in the target corpus.
Our aim is to estimate a topic model for the given target corpus.
Figure~\ref{fig:method} shows our framework.

The neural networks in our model take support data as input, 
and output topic model priors.
First, our model obtains a corpus representation by a permutation invariant neural network~\cite{zaheer2017deep}
from the support data.
The corpus representation contains the property of the support data.
Next, Dirichlet parameters, which define the prior distribution of topic model parameters,
are obtained by neural networks using the corpus representation and support data.
By using the corpus representation, the obtained Dirichlet parameters become corpus-specific.
Then, topic model parameters are obtained by maximizing the posterior probability given the support data and the priors
with the expectation-maximization (EM) algorithm~\cite{dempster1977maximum}.
Each step in the EM algorithm
can be seen as a layer of a neural network in our model.
Since our model is differentiable, including the EM steps, the loss can be backpropagated through our model.
Therefore, we can train neural networks such that
the obtained topic model achieves high generalization performance when learned with the EM algorithm.
Although the EM algorithm has been used for estimating topic model parameters~\cite{hofmann1999probabilistic},
it has not been used as layers of a neural network.

The main contributions of this paper
are as follows:
\begin{enumerate}
\item We propose a method that can learn topic models from just a few documents.
\item We backpropagate the loss through the EM algorithm to train neural networks that output parameters of probability distributions.
\item We demonstrate that the proposed method
performs better than existing methods using real-world text document datasets.
\end{enumerate}

\section{Related work}
\label{sec:related}

Most existing few-shot learning methods were designed for supervised learning~\cite{schmidhuber:1987:srl,bengio1991learning,ravi2016optimization,andrychowicz2016learning,vinyals2016matching,snell2017prototypical,bartunov2018few,finn2017model,li2017meta,kimbayesian,finn2018probabilistic,rusu2018meta,yao2019hierarchically,garnelo2018conditional,kim2019attentive,hewitt2018variational,bornschein2017variational,rezende2016one,tang2019,narwariya2020meta,xie2019meta,lake2019compositional,iwata2020meta}.
Some unsupervised few-shot learning methods have been proposed,
such as clustering~\cite{hsu2018unsupervised,kim2019meta}
and density estimation~\cite{edwards2016towards,reed2018few},
but they are not for topic modeling.
Model-agnostic meta-learning (MAML)~\cite{finn2017model} is related to the proposed method
since both learn neural networks such that they perform better when fine-tuned with the support data.
MAML uses a gradient descent method for fine-tuning,
where the gradient of the gradient is needed for backpropagation through the gradient descent steps,
which can be costly in terms of memory~\cite{bertinetto2018meta}.
On the other hand, the proposed method uses the gradient of the EM algorithm,
where each EM step is obtained by simple calculation
in a closed form for topic models.
In addition, the EM algorithm does not use any hyperparameters,
whereas gradient descent methods use hyperparameters,
such as learning rate.
The proposed method is also related to encoder-decoder style meta-learning methods~\cite{xu2019metafun} like conditional neural processes~\cite{garnelo2018conditional}, where a task (corpus) representation is obtained by encoder neural networks from support data. However, they are different after the encoding. The conditional neural processes directly output labels by decoder neural networks. In contrast, neural networks with the proposed method output topic model priors, which are used by the EM algorithm to estimate the topic model parameters.
Several topic models for short texts~\cite{lin2014dual,zuo2016word},
continual lifelong learning~\cite{gupta2020neural},
and transfer learning~\cite{faisal2012sparse,kang2012transfer,acharya2013using,hu2015transfer} have been proposed.
However, they fail to handle document paucity.

\section{Proposed method}
\label{sec:proposed}

\subsection{Problem formulation}

Suppose that we are given word frequency vectors of documents in
$D$ corpora $\{\vec{X}_{d}\}_{d=1}^{D}$ in the training phase,
where $\vec{X}_{d}=\{\vec{x}_{dn}\}_{n=1}^{N_{d}}$
is a set of documents in the $d$th corpus,
$N_{d}$ is the number of documents,
$\vec{x}_{dn}=(x_{dnj})_{j=1}^{J}$ is the word frequency vector of the $n$th document,
$x_{dnj}$ is the word frequency of the $j$th vocabulary term,
and $J$ is the number of vocabulary terms.
The number of vocabulary terms is the same for all corpora,
but the number of documents can be different across corpora.
In each task, documents are assumed to be generated from a corpus-specific probability distribution.
In the test phase,
we are given a small number of documents $\vec{X}_{*}=\{\vec{x}_{*n}\}_{n=1}^{N_{*}}$ in the target corpus,
which is different from but related to the training corpora.
Our aim is to estimate the topics of the target corpus.
Table~\ref{tab:notation} summarizes our notation. 

\begin{table}[t!]
\centering
\caption{Our notation}
\label{tab:notation}
\begin{tabular}{ll}
\hline
Symbol & Description \\
\hline
$\vec{X}_{d}$ & word frequency vectors of the $d$th corpus\\
$K$ & number of topics\\
$N$ & number of documents\\
$J$ & number of vocabulary terms\\
$D$ & number of corpora\\
$T$ & number of EM steps\\
$\theta_{nk}$ & probability that document $n$ includes topic $k$\\
$\phi_{kj}$ & probability that term $j$ occurs in topic $k$\\
$\vec{r}$ & corpus representation\\
$\bm{\alpha}_{n}$ & Dirichlet parameters for the prior of $\bm{\theta}_{n}$\\
$\bm{\beta}_{k}$ & Dirichlet parameters for the prior of $\bm{\phi}_{k}$\\
$\vec{X}$ & support data\\
$\vec{X}'$ & query data\\
$\bm{\Psi}$ & neural network parameters\\
\hline
\end{tabular}
\end{table}

\subsection{Topic model}

With the proposed method,
documents in a corpus are assumed to be generated according to a topic model~\cite{blei2003latent}.
A topic model consists of topic proportion vectors $\bm{\Theta}=(\bm{\theta}_{n})_{n=1}^{N}$ and
word probability vectors $\bm{\Phi}=(\bm{\phi}_{k})_{k=1}^{K}$,
where $K$ is the number of topics,
$\bm{\theta}_{n}=(\theta_{nk})_{k=1}^{K}$,
$\theta_{nk}$ is the probability that the $n$th document has the $k$th topic,
$\sum_{k=1}^{K}\theta_{nk}=1$, $\theta_{nk}\geq 0$,
$\bm{\phi}_{k}=(\phi_{kj})_{j=1}^{J}$, 
$\phi_{kj}$ is the probability that the $j$th term occurs in the $k$th topic,
$\sum_{j=1}^{J}\phi_{kj}=1$, and $\phi_{kj}\geq 0$.
Here, we omit corpus index $d$ for simplicity.

With the topic model,
the probability of word frequency vectors in a corpus, $\vec{X}=(\vec{x}_{n})_{n=1}^{N}$, is given by
\begin{align}
  p(\vec{X}|\bm{\Theta},\bm{\Phi})=\prod_{n=1}^{N}\prod_{j=1}^{J}\left(\sum_{k=1}^{K}\theta_{nk}\phi_{kj}\right)^{x_{nj}}.
  \label{eq:px}
\end{align}
The topic proportion vector is assumed to follow a Dirichlet distribution,
\begin{align}
  p(\bm{\theta}_{n}|\bm{\alpha}_{n})
  = \frac{\Gamma(\sum_{k=1}^{K}\alpha_{nk}+K)}{\prod_{k=1}^{K}\Gamma(\alpha_{nk}+1)}
  \prod_{k=1}^{K}\theta_{nk}^{\alpha_{nk}},
  \label{eq:ptheta}
\end{align}
where $\bm{\alpha}_{n}=(\alpha_{nk})_{k=1}^{K}$ are the Dirichlet parameters,
and $\Gamma(\cdot)$ is the gamma function.
The Dirichlet distribution is the conjugate prior for topic proportions $\bm{\theta}_{n}$.
Note that we use an asymmetric Dirichlet prior, so the parameters can differ across topics,
i.e., $\alpha_{nk}\neq\alpha_{nk'}$.
Similarly, the word probability vector is assumed to follow a Dirichlet distribution,
\begin{align}
  p(\bm{\phi}_{k}|\bm{\beta}_{k})
  = \frac{\Gamma(\sum_{j=1}^{J}\beta_{kj}+J)}{\prod_{j=1}^{J}\Gamma(\beta_{kj}+1)}
  \prod_{j=1}^{J}\phi_{kj}^{\beta_{kj}},
  \label{eq:pphi}
\end{align}
where $\bm{\beta}_{k}=(\beta_{kj})_{j=1}^{J}$ are the Dirichlet parameters.
The Dirichlet distribution is the conjugate prior for word probability vector $\bm{\phi}_{k}$.

\subsection{Topic model generator}

Our model generates topic model parameters, $\bm{\Theta}$ and $\bm{\Phi}$, given  support data,
$\vec{X}=(\vec{x}_{n})_{n=1}^{N}\in\mathbb{Z}_{\geq 0}^{N\times J}$, which is a matrix consisting of word frequency vectors on a small number of documents.

First, we calculate corpus representation $\vec{r}\in\mathbb{R}^{M}$ using support data $\vec{X}$
by a permutation invariant neural network~\cite{zaheer2017deep} as follows,
\begin{align}
  \vec{r} = g_{\mathrm{R}}\left(\frac{1}{N}\sum_{n=1}^{N}f_{\mathrm{R}}(\vec{x}_{n})\right),
  \label{eq:r}
\end{align}
where $f_{\mathrm{R}}$ and $g_{\mathrm{R}}$ are feed-forward neural networks.
We use permutation invariant neural networks since the corpus representation should not depend on
the order of the documents in the support data.
In addition, we can obtain a corpus representation even when the number of documents $N$ in the support data is changed.

Second,
we obtain Dirichlet parameters $\bm{\alpha}_{n}$ for the prior of topic proportion vector $\bm{\theta}_{n}$
using corpus representation $\vec{r}$ and word frequency vector $\vec{x}_{n}$ of each document as follows,
\begin{align}
  \bm{\alpha}_{n} = f_{\mathrm{A}}([\vec{x}_{n},\vec{r}]),
  \label{eq:alpha}
\end{align}
where $f_{\mathrm{A}}$ is a feed-forward neural network,
and
$[\cdot,\cdot]$ is vector concatenation.
Since the corpus representation is used, the Dirichlet parameters contain properties of the given corpus.

Third,
we obtain Dirichlet parameters $\bm{\beta}_{k}$ for the prior of word probability vector $\bm{\phi}_{k}$
using corpus representation $\vec{r}$,
Dirichlet parameters for topic proportions $\bm{\alpha}_{\cdot k}=(\alpha_{nk})_{n=1}^{N}\in\mathbb{R}^{N}$,
and support data $\vec{X}$ as follows,
\begin{align}
  \bm{\beta}_{k} = f_{\mathrm{B}}([\vec{X}^{\top}\bm{\alpha}_{\cdot k},\vec{r}]),
  \label{eq:beta}  
\end{align}
where $f_{\mathrm{B}}$ is a feed-forward neural network,
and $\top$ is transpose.
By multiplying $\vec{X}$ and $\bm{\alpha}_{\cdot k}$,
the information on the $k$th topic of the given corpus can be extracted
considering the Dirichlet parameters for topic proportions on the $k$th topic.
In addition, since $\vec{X}^{\top}\bm{\alpha}_{\cdot k}$ is a vector with size $J$,
the input layer size of $f_{\mathrm{B}}$ does not depend on the number of documents $N$,
so our model can handle data with different numbers of documents.

Fourth,
we estimate $\bm{\Theta}$ and $\bm{\Phi}$ by maximizing the posterior probability as follows,
\begin{align}
        \hat{\bm{\Theta}},\hat{\bm{\Phi}}&=\arg\max_{\bm{\Theta},\bm{\Phi}} p(\vec{\Theta},\vec{\Phi}|\vec{X})
        \nonumber\\
        &=\arg\max_{\bm{\Theta},\bm{\Phi}} \bigl[ \log p(\vec{X}|\vec{\Theta},\vec{\Phi})+\log p(\vec{\Theta},\vec{\Phi})\bigr],
  \label{eq:L}  
\end{align}
where we used the Bayes rule in the second equality.
Let $\mathcal{L}=\log p(\vec{X}|\vec{\Theta},\vec{\Phi})+\log p(\vec{\Theta},\vec{\Phi})+C$ be the objective function, where $C$ is the constant term.
We find its local optimum solution by the EM algorithm~\cite{dempster1977maximum},
where the parameters are updated by maximizing the following lower bound of the objective function,
\begin{align}
\mathcal{L}&=
\sum_{n=1}^{N}\sum_{j=1}^{J}x_{nj}\log\sum_{k=1}^{K}\theta_{nk}\phi_{kj}
\nonumber\\        
&+\sum_{n=1}^{N}\sum_{k=1}^{K}\alpha_{nk}\log\theta_{nk}
+\sum_{k=1}^{K}\sum_{j=1}^{J}\beta_{kj}\log\phi_{kj}
\nonumber\\           
&\geq
\sum_{n=1}^{N}\sum_{j=1}^{J}x_{nj}\sum_{k=1}^{K}\gamma_{njk}\log\frac{\theta_{nk}\phi_{kj}}{\gamma_{njk}}
\nonumber\\           
&+\sum_{n=1}^{N}\sum_{k=1}^{K}\alpha_{nk}\log\theta_{nk}
+\sum_{k=1}^{K}\sum_{j=1}^{J}\beta_{kj}\log\phi_{kj} \equiv Q,
\label{eq:Q}
\end{align}
where
we used Eqs.~(\ref{eq:px},\ref{eq:ptheta},\ref{eq:pphi}) in the first equality,
we used Jensen's inequality,
$\gamma_{njk}$ is the responsibility that represents
the probability that the $j$th term in the $n$th document has the $k$th topic,
$\gamma_{njk}\geq0$, and $\sum_{k=1}^{K}\gamma_{njk}=1$.
We initialize the parameters by the mode of the Dirichlet distributions using
Dirichlet parameters in Eqs.~(\ref{eq:alpha},\ref{eq:beta}) as follows,
\begin{align}
  \theta_{nk}=\frac{\alpha_{nk}}{\sum_{k'=1}^{K}\alpha_{nk'}},
  \label{eq:theta0}
\end{align}
\begin{align}
  \phi_{kj}=\frac{\beta_{jk}}{\sum_{j'=1}^{J}\beta_{j'k}}.
  \label{eq:phi0}
\end{align}
With the E-step,
the responsibility is calculated by
\begin{align}
  \gamma_{njk}=\frac{\theta_{nk}\phi_{kj}}{\sum_{k'=1}^{K}\theta_{nk'}\phi_{k'j}},
  \label{eq:gamma}    
\end{align}
which
is obtained analytically by maximizing the lower bound $Q$ in Eq.~(\ref{eq:Q}) with respect to $\gamma_{njk}$.
With the M-step,
the parameters are updated using the responsibility by
\begin{align}
  \theta_{nk}=\frac{\sum_{j=1}^{J}x_{nj}\gamma_{njk}+\alpha_{nk}}{\sum_{k'=1}^{K}(\sum_{j=1}^{J}x_{nj}\gamma_{njk'}+\alpha_{nk'})},
  \label{eq:theta}     
\end{align}
\begin{align}
  \phi_{kj}=\frac{\sum_{n=1}^{N}x_{nj}\gamma_{njk}+\beta_{jk}}{\sum_{j'=1}^{J}(\sum_{n=1}^{N}x_{nj'}\gamma_{nj'k}+\beta_{j'k})}.
  \label{eq:phi}     
\end{align}
Eqs.~(\ref{eq:theta},\ref{eq:phi}) are obtained analytically
by maximizing the lower bound $Q$ in Eq.~(\ref{eq:Q}) with respect to
$\theta_{nk}$ and $\phi_{kj}$, respectively.
The E- and M-steps are iterated $T$ times or until convergence, which yields estimates of topic model
parameters $\hat{\bm{\Theta}}$ and $\hat{\bm{\Phi}}$.
The EM algorithm is guaranteed to monotonically increase the posterior probability at each step until it reaches a local maximum.

\begin{figure}[t!]
  \centering
  \includegraphics[width=19em]{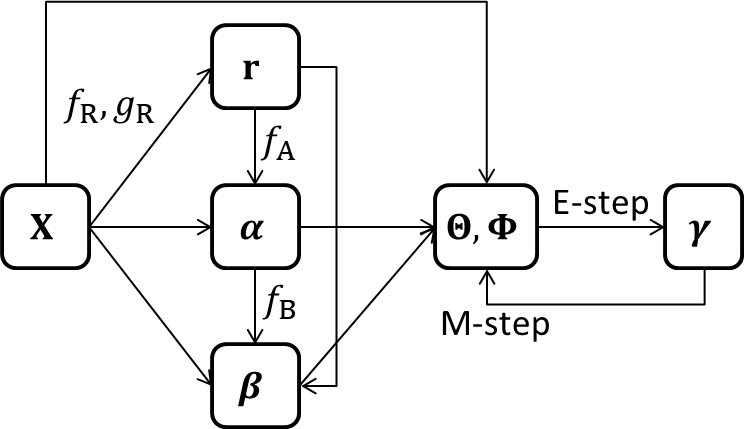}
  \caption{Our model that takes support data $\vec{X}$ as input, and outputs topic model parameters $\bm{\Theta}$ and $\bm{\Phi}$ using feed-forward neural networks $f_{\mathrm{R}}$, $g_{\mathrm{R}}$, $f_{\mathrm{A}}$, $f_{\mathrm{B}}$, and EM steps.}
  \label{fig:model}
\end{figure}

Our model including the E- and M-steps can be seen as a single neural network
that takes support data $\vec{X}$ as input and outputs
estimated topic model parameters $\hat{\bm{\Theta}}$ and $\hat{\bm{\Phi}}$.
We call Eqs.~(\ref{eq:gamma}--\ref{eq:phi}) the EM layers.
Since the EM layers are differentiable,
we can backpropagate the loss through our model.
Algorithm~\ref{alg:model} and Figure~\ref{fig:model} show the procedures of our model.
Note that although
we use iterations over documents $n$, topics $k$, and vocabulary terms $j$ for clarity in Algorithm~\ref{alg:model},
we can efficiently calculate them by matrix calculations.
The EM algorithm described in this section
can be straightforwardly replaced by variational Bayesian inference, which is also differentiable.

\begin{algorithm}[t!]
  \caption{Our model.}
  \label{alg:model}
  \begin{algorithmic}[1]
    \renewcommand{\algorithmicrequire}{\textbf{Input:}}
    \renewcommand{\algorithmicensure}{\textbf{Output:}}
    \REQUIRE{Suppot data $\vec{X}$, and number of EM steps $T$}
    \ENSURE{Topic model parameters $\hat{\bm{\Theta}}$, $\hat{\bm{\Phi}}$}
    \STATE Calculate corpus representation $\vec{r}$ by Eq.~(\ref{eq:r})
    \\ \vspace{1em} \textit{\#Calculate topic model priors}
    \FOR{each document $n:=1$ to $N$} 
    \STATE Calculate topic proportion Dirichlet prior parameters $\bm{\alpha}_{n}$ by Eq.~(\ref{eq:alpha})
    \ENDFOR
    \FOR{each topic $k:=1$ to $K$}
    \STATE Calculate word distribution Dirichlet prior parameters $\bm{\beta}_{k}$ by Eq.~(\ref{eq:beta})
    \ENDFOR
    \\ \vspace{1em} \textit{\#Initialize topic model parameters}
    \FOR{each document $n:=1$ to $N$}
    \FOR{each topic $k:=1$ to $K$}
    \STATE Initialize topic proportions $\theta_{nk}$ by Eq.~(\ref{eq:theta0})
    \ENDFOR
    \ENDFOR
    \FOR{each topic $k:=1$ to $K$}
    \FOR{each vocabulary term $j:=1$ to $J$}
    \STATE Initialize word distribution $\phi_{kj}$ by Eq.~(\ref{eq:phi0})
    \ENDFOR
    \ENDFOR
    \\ \vspace{1em} \textit{\#Estimate topic model parameters}
    \FOR{each EM step $t:=1$ to $T$}
    \FOR{each document $n:=1$ to $N$}
    \FOR{each vocabulary term $j:=1$ to $J$}
    \FOR{each topic $k:=1$ to $K$}
    \STATE Calculate responsibility $\gamma_{njk}$ by Eq.~(\ref{eq:gamma})
    \ENDFOR
    \ENDFOR
    \ENDFOR
    \FOR{each document $n:=1$ to $N$}
    \FOR{each topic $k:=1$ to $K$}
    \STATE Update topic proportions $\theta_{nk}$ by Eq.~(\ref{eq:theta})
    \ENDFOR
    \ENDFOR
    \FOR{each topic $k:=1$ to $K$}
    \FOR{each vocabulary term $j:=1$ to $J$}
    \STATE Update word distribution $\phi_{kj}$ by Eq.~(\ref{eq:phi})
    \ENDFOR
    \ENDFOR
    \ENDFOR
  \end{algorithmic}
\end{algorithm}

\subsection{Training}

The parameters of neural networks in our model, $f_{\mathrm{R}}$, $g_{\mathrm{R}}$, $f_{\mathrm{A}}$, and $f_{\mathrm{B}}$,
are trained by maximizing the expected test performance using an episodic training framework~\cite{ravi2016optimization,santoro2016meta,snell2017prototypical,finn2017model,li2019episodic},
where the test phase is simulated by using the training corpora.

For each epoch, we randomly select $N$ documents $\bar{\vec{X}}$
from a randomly selected training corpus $\vec{X}_{d}$,
where $\bar{\vec{X}}\subset\vec{X}_{d}$.
The selected documents $\bar{\vec{X}}$ are randomly split into 
support data $\vec{X}$ and query data $\vec{X}'$,
where $\bar{\vec{X}}=\vec{X}+\vec{X}'$.
The support data are used for obtaining topic model parameters,
$\hat{\bm{\Theta}}$ and $\hat{\bm{\Phi}}$, by inputting them to our model.
The query data are used for evaluating the expected test performance.
Algorithm~\ref{alg:split} shows the splitting procedure of the data into support and query data,
where each word in the given data is randomly included in the support data with probability $R$,
and in the query data with probability $1-R$. 

\begin{algorithm}[t!]
  \caption{$\mathrm{Split}(\bar{\vec{X}},R)$: Splitting procedure of word frequency vectors $\bar{\vec{X}}\in\mathbb{Z}_{\geq 0}^{N\times J}$
  into support data $\vec{X}\in\mathbb{Z}_{\geq 0}^{N\times J}$ and query data $\vec{X}'\in\mathbb{Z}_{\geq 0}^{N\times J}$.}
  \label{alg:split}
  \begin{algorithmic}[1]
    \renewcommand{\algorithmicrequire}{\textbf{Input:}}
    \renewcommand{\algorithmicensure}{\textbf{Output:}}
    \REQUIRE{Word frequency vectors $\bar{\vec{X}}$, support rate $R$}
    \ENSURE{Support data $\vec{X}$, query data $\vec{X}'$}
    \STATE Initialize support and query data, $\vec{X}:=\vec{0}$, $\vec{X}':=\vec{0}$
    \FOR{each document $n:=1$ to $N$}
    \FOR{each vocabulary term $j:=1$ to $J$}
    \FOR{each word count $i:=1$ to $\bar{x}_{nj}$}
    \STATE Randomly generate a binary value, \\ $r\sim\mathrm{Bernoulli}(R)$
    \IF{$r=1$}
    \STATE Include the word in support data, \\
    $x_{nj}:=x_{nj}+1$
    \ELSE
    \STATE Include the word in query data, \\
    $x'_{nj}:=x'_{nj}+1$
    \ENDIF
    \ENDFOR
    \ENDFOR
    \ENDFOR
  \end{algorithmic}
\end{algorithm}

The objective function to be maximized is the following expected test log likelihood,
\begin{align}
  \hat{\bm{\Psi}}=\arg\max_{\bm{\Psi}}\mathbb{E}_{d}[\mathbb{E}_{(\vec{X},\vec{X}')\sim\vec{X}_{d}} [
      L(\vec{X}'|\hat{\bm{\Theta}}(\vec{X};\bm{\Psi}),\hat{\bm{\Phi}}(\vec{X};\bm{\Psi}))]],
\end{align}
where $\bm{\Psi}$ is the set of parameters of neural networks in our model,
$\mathbb{E}$ is the expectation,
$(\vec{X},\vec{X}')\sim\vec{X}_{d}$ represents a random sampling of support $\vec{X}$ and query data $\vec{X}'$ from $\vec{X}_{d}$,
\begin{align}
  L(\vec{X}'|\bm{\Theta},\bm{\Phi})
  =\sum_{n=1}^{N}\sum_{j=1}^{J}x_{nj}'\log\sum_{k=1}^{K}\theta_{nk}\phi_{kj},
\end{align}
is the test log likelihood of topic model parameters given $\vec{X}'$,
and $\hat{\bm{\Theta}}(\vec{X};\bm{\Psi})$ and $\hat{\bm{\Phi}}(\vec{X};\bm{\Psi})$
are topic model parameters that are obtained
by our model with neural network parameter $\bm{\Psi}$
from support data $\vec{X}$.
Algorithm~\ref{alg:train} shows the training procedures of our model.

\begin{algorithm}[t!]
  \caption{Training procedure of our model.}
  \label{alg:train}
  \begin{algorithmic}[1]
    \renewcommand{\algorithmicrequire}{\textbf{Input:}}
    \renewcommand{\algorithmicensure}{\textbf{Output:}}
    \REQUIRE{Training datasets $\{\vec{X}_{d}\}_{d=1}^{D}$, support size $N$,
      support rate $R$}
    \ENSURE{Trained neural network parameters $\bm{\Psi}$}
    \STATE Initialize neural network parameters $\bm{\Psi}$
    \WHILE{End condition is satisfied}
    \STATE Uniform randomly select corpus index $d$ from $\{1,\cdots,D\}$
    \STATE Randomly select $N$ documents $\bar{\vec{X}}$ from the selected corpus $\vec{X}_{d}$
    \STATE Randomly split the selected documents into support and query data by Algorithm~\ref{alg:split}, $\vec{X},\vec{X}':= \mathrm{Split}(\bar{\vec{X}},R)$
    \STATE Obtain topic model parameters, $\hat{\bm{\Theta}}(\vec{X};\bm{\Psi})$ and $\hat{\bm{\Phi}}(\vec{X};\bm{\Psi})$, by our model in Algorithm~\ref{alg:model} 
    \STATE Calculate loss $-L(\vec{X}'|\hat{\bm{\Theta}}(\vec{X};\bm{\Psi}),\hat{\bm{\Phi}}(\vec{X};\bm{\Psi}))$ and its gradients
    \STATE Update neural network parameters $\bm{\Psi}$ using the loss and its gradient by a stochastic gradient method
    \ENDWHILE
  \end{algorithmic}
\end{algorithm}


\section{Experiments}
\label{sec:experiments}

\subsection{Data}

We evaluated the proposed method
using three datasets: 20News, Digg, and NeurIPS.
The 20News data were obtained from the 20 Newsgroups corpus~\cite{lang1995newsweeder}.
The Digg data were obtained from Digg, which is a social news service.
The NeurIPS data were obtained from papers in Conferences on Neural Information Processing Systems from 2001 to 2003~\footnote{Available at \url{http://robotics.stanford.edu/~gal/data.html}}.
From each data set,
we omitted documents that contained less than 30 vocabulary terms, and vocabulary terms that appeared in less than 30 documents.
The 20News data yielded 14,366 documents, 7,364 vocabulary terms, and 20 categories.
The Digg data yielded 3,859 documents, 776 vocabulary terms, and 23 categories.
The NeurIPS data yielded 592 documents, 2,345 vocabulary terms, and 13 categories.
For each data set,
we used documents in a category as the target corpus,
documents in three categories as the validation data,
and documents in the remaining categories as the training data,
where a corpus consisted of documents in a category.
With the target corpus,
20\% of the words were held out and used for evaluations,
and the remaining words were used as the support data in the test phase.
The number of documents in the target corpus was three,
and we trained models with support size $N=3$, and support rate $R=0.8$.
For each target corpus, we conducted ten experiments with different training and validation splits,
and different support and query splits in the target corpus.

\subsection{Neural network architecture}

For neural networks, $f_{\mathrm{R}}$, $g_{\mathrm{R}}$, $f_{\mathrm{A}}$, and $f_{\mathrm{B}}$,
we used three-layered feed-forward neural networks with 256 hidden units.
The input and output unit sizes of $f_{\mathrm{R}}$ were
$J$ and 256,
those of $g_{\mathrm{R}}$ were 256 and 256,
those of $f_{\mathrm{A}}$ were $J$+256 and $K$,
and
those of $f_{\mathrm{B}}$ were $J$+256 and $J$, respectively.
For the activation function, we used rectified linear unit $\mathrm{ReLU}(x)=\max(0,x)$.
We used softplus functions $\log(1+\exp(x))$ at the end of $f_{\mathrm{A}}$ and $f_{\mathrm{B}}$
so that $\bm{\alpha}_{n}$ and $\bm{\beta}_{k}$ were non-negative.
The number of EM steps was $T=10$.
We optimized using Adam~\cite{kingma2014adam}
with learning rate $10^{-3}$,
and dropout rate $10^{-1}$~\cite{srivastava2014dropout}.
The validation data were used for early stopping,
where the maximum number of training epochs was 1,000.
Our implementation was based on PyTorch~\cite{paszke2017automatic}.

\subsection{Comparing methods}

We compared the proposed method with
the following 11 topic model parameter estimation methods:
LDAind, LDAall, MAML, NN, NN-R, NN-F, NN-RF, NN-E, Dir, Dir-F, and Dir-E.
LDA was the latent Dirichlet allocation inferred by collapsed Gibbs sampling~\cite{griffiths2004finding},
where the Dirichlet parameters were estimated by the fixed-point iteration method~\cite{minka2000estimating}.
LDAind was trained individually using the target support data,
while LDAall was trained using all training data.
MAML optimized the Dirichlet parameters
using model-agnostic meta-learning~\cite{finn2017model}
by maximizing the expected test log-likelihood.
For the inner loop with MAML, we used Adam with five steps.
The NN method obtained the Dirichlet parameters using neural networks (NN)
without corpus representation $\vec{r}$ by
$\bm{\alpha}_{n}=f_{\mathrm{A}}(\vec{x}_{n})$ instead of Eq.~(\ref{eq:alpha}),
and
$\bm{\beta}_{k}=f_{\mathrm{B}}(\vec{X}^{\top}\bm{\alpha}_{\cdot k})$ instead of Eq.~(\ref{eq:beta}).
The topic model parameters were estimated by the Dirichlet parameters with Eqs.~(\ref{eq:theta0},\ref{eq:phi0})
without the EM layers.
The NN-R method was the same as NN except that that it used corpus representation $\vec{r}$ by Eqs.~(\ref{eq:alpha}, \ref{eq:beta})
to obtain the Dirichlet parameters.
The NN-F method was NN with fine-tuning, where the target support data were used for the fine-tuning based on the EM algorithm in the test phase.
The NN-RF method was NN-R with fine-tuning.
The NN-E method was NN with the EM layers.
The difference between NN-E and the proposed method was that NN-E did not use the corpus representation.
The Dir method optimized the Dirichlet parameters directly without neural networks by maximizing the expected test log-likelihood.
The topic model parameters were estimated with Eqs.~(\ref{eq:theta0},\ref{eq:phi0}) without the EM layers.
The Dir-F method was Dir with fine-tuning in the test phase.
The Dir-E method was Dir with the EM layers.

\subsection{Results}

Table~\ref{tab:perp} shows the test perplexity results,
where lower test perplexity indicates
better performance of the estimated topic model.
The proposed method achieved the best perplexity for all data sets.
Since the LDAind method could not use the information on related corpora, its perplexity was high.
Although the LDAall method was trained using data in related corpora,
the estimated topic model was not corpus specific.
Therefore, the LDAall method yielded worse perplexity than the proposed method.
The perplexity of the NN method was lower than that of the Dir method.
The Dir method estimated the Dirichlet parameters that were shared across all corpora.
On the other hand, the NN method estimated the Dirichlet parameters that depended on the given support data using neural networks.
This result indicates that using neural networks to obtain the corpus dependent Dirichlet parameters is effective.
The NN-R method had better performance than the NN method for 20News and Digg data sets,
but worse for the NeurIPS data set.
The use of the corpus representation can degrade the performance if the EM layers are omitted.
Since the proposed method directly maximizes the likelihood given the support data by the EM layers, it mitigates 
the adverse effect of the estimated corpus representation,
which improves the performance.
The NN-E method achieved lower perplexity than the NN method due to its EM layers; the neural networks were trained such that fine-tuning yielded better performance.
In contrast, the NN-F method could not improve the performance from the NN method for the 20News and Digg data sets
since the neural networks were trained without consideration of fine-tuning.
MAML yielded worse perplexity than the Dir-E method for the 20News and NeurIPS data sets.
The difference between the MAML method and Dir-E methods was
that MAML fine-tuned the neural networks by stochastic gradient methods, whereas the Dir-E method fine-tuned them by the EM algorithm.

\begin{table}[t!]
  \centering
  \caption{Average test perplexity on target corpora and the standard error for each dataset. Values in bold typeface are not statistically different at 5\% level from the best performing method in each dataset according to paired t-tests.}
  \label{tab:perp}
  {\tabcolsep=0.4em
    \begin{tabular}{lccc}
      \hline
 &  20News  &       Digg  &       NeurIPS  \\
 \hline
Ours & {\bf 2785.8 $\pm$ 67.0}  &  {\bf 556.1 $\pm$ 9.7}  &  {\bf 606.7 $\pm$ 6.7}  \\
LDAind & 3239.1 $\pm$ 88.0  &  613.4 $\pm$ 13.2  &  636.2 $\pm$ 9.0 \\
LDAall & 3542.8 $\pm$ 98.8  &  631.7 $\pm$ 27.6  &  926.8 $\pm$ 9.9\\
MAML & 4748.2 $\pm$ 56.9  &  645.9 $\pm$ 5.7  &  1404.6 $\pm$ 9.1 \\
NN & 2964.4 $\pm$ 68.7  &  584.9 $\pm$ 9.7  &  746.6 $\pm$ 8.2 \\
NN-R & 2944.2 $\pm$ 68.9  &  578.4 $\pm$ 9.8  &  827.7 $\pm$ 9.4 \\
NN-F & 4560.7 $\pm$ 131.6  &  620.3 $\pm$ 12.6  &  742.3 $\pm$ 9.2 \\
NN-RF & 3994.2 $\pm$ 115.4  &  589.7 $\pm$ 11.1  &  835.0 $\pm$ 9.2 \\
NN-E & 2831.7 $\pm$ 68.4  &  568.8 $\pm$ 9.9  &  667.3 $\pm$ 6.8 \\
Dir & 6142.0 $\pm$ 19.9  &  699.6 $\pm$ 2.5  &  1725.6 $\pm$ 4.8 \\
Dir-F & 5468.4 $\pm$ 56.4  &  688.8 $\pm$ 2.6  &  988.4 $\pm$ 9.9 \\
Dir-E & 4629.2 $\pm$ 70.1  &  653.1 $\pm$ 4.7  &  873.7 $\pm$ 7.8 \\
\hline
  \end{tabular}}
\end{table}

\begin{figure*}[t!]
  \centering
      {\tabcolsep=0.8em
        \begin{tabular}{ccc}
          \includegraphics[width=12.5em]{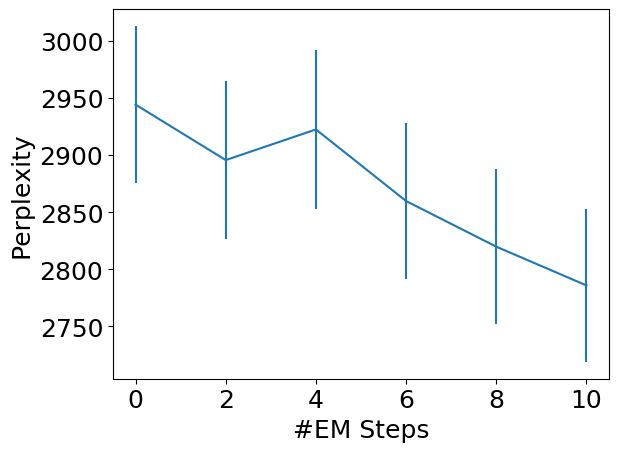} &
          \includegraphics[width=12.5em]{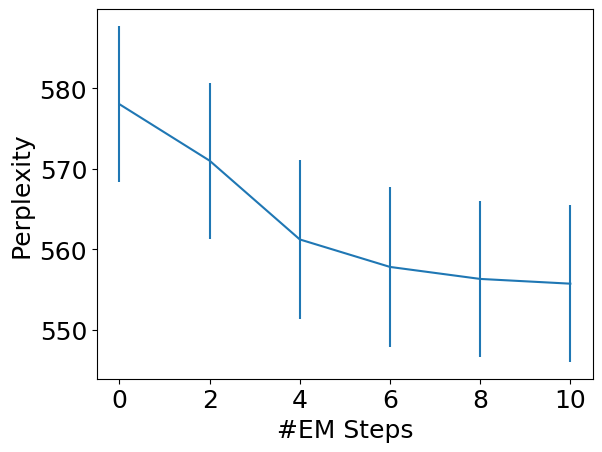} &
          \includegraphics[width=12.5em]{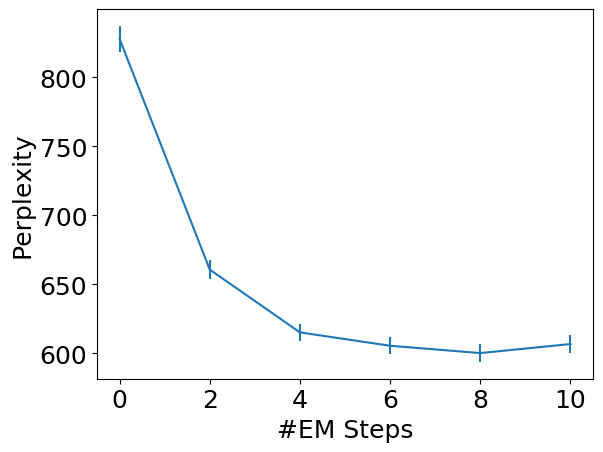} \\
          (a) 20News & (b) Digg & (c) NeurIPS \\
      \end{tabular}}
      \caption{Average test perplexity with different numbers of EM steps in the proposed method. The bars show the standard error.}
      \label{fig:em}
\end{figure*}

\begin{figure*}[t!]
  \centering
      {\tabcolsep=0.8em
        \begin{tabular}{ccc}
          \includegraphics[width=12.5em]{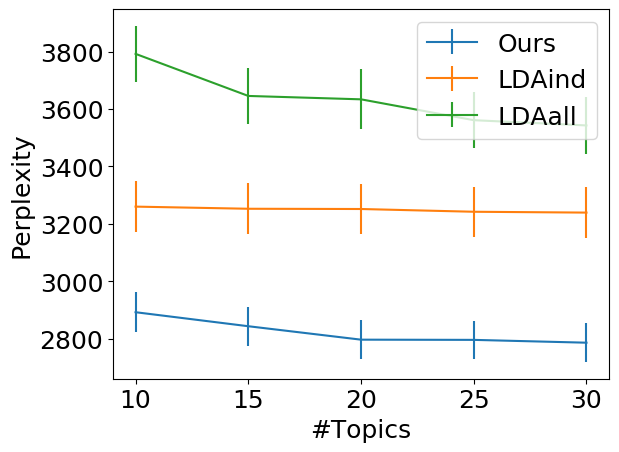} &
          \includegraphics[width=12.5em]{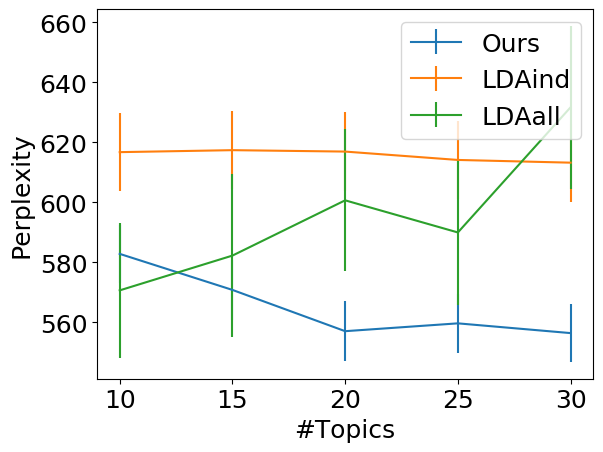} &
          \includegraphics[width=12.5em]{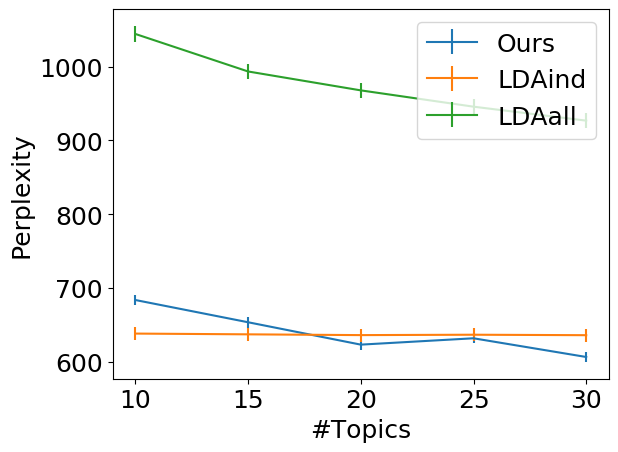} \\
          (a) 20News & (b) Digg & (c) NeurIPS \\
      \end{tabular}}
      \caption{Average test perplexity with different numbers of topics in the proposed method. The bars show the standard error.}
      \label{fig:topic}
\end{figure*}

\begin{table}[t!]
  \centering
  \caption{Training computational time in seconds.}
  \label{tab:time}
  {\tabcolsep=0.5em
    \begin{tabular}{lrrr}
      \hline
& 20News  &       Digg  &       NeurIPS \\
 \hline 
Ours & 10890.0 &  1314.8  &  1750.7 \\
LDAind & 25.7  &  3.0   &  18.5 \\
LDAall & 10455.4 & 435.4 & 3646.0 \\
MAML & 5612.3  &  1060.6 &  1294.3 \\
NN & 2688.1 &  387.2   &  641.6 \\
NN-R & 3578.4 &  505.1  &  840.2 \\
NN-E & 10062.9 &  1196.8  &  1600.2 \\
Dir & 583.6 &  117.8  &  279.0 \\
Dir-E & 8282.3 &  1150.5  &  1439.2 \\
 \hline
  \end{tabular}}
\end{table}

\begin{table*}[t!]
\centering
\caption{Estimated topics in three corpora with the 20News data set. Vocabulary terms that have high word probability $\phi_{kj}$ for three topics are shown.}
\label{tab:topic}
\begin{tabular}{ll}
\hline
Corpus & sci.space\\
\hline
Topic1 & mission missions addition orbit space months time program years power\\
Topic2 & program years months time spacecraft solar question interaction propulsion mass\\
Topic3 & power earth electricity field interaction propulsion zoology utzoo resembles zoo\\
\hline
Corpus & talk.politics.mideast\\
\hline
Topic1 & arabs peace arab land palestinians make culture palestine talks account\\
Topic2 & negotiating tight moves suspended chances involving dramatic december faced evidently\\
Topic3 & turks turkish negotiating tight moves suspended chances involving dramatic december\\
\hline
Corpus & comp.sys.mac.hardware\\
\hline
Topic1 & clock plastic mhz power tower drive remove speed socket case\\
Topic2 & scsi tim questions powerbook told expensive bet town lineup jose\\
Topic3 & place mac side powerbook software drive expensive bet town lineup\\
\hline
\end{tabular}
\end{table*}

Figure~\ref{fig:em} shows the test perplexity of the proposed method with different numbers of EM steps.
As the number of the EM steps increased, the perplexity decreased.
Figure~\ref{fig:topic} shows the test perplexity with different numbers of topics attained by the proposed method, LDAind, and LDAall.
Particularly when the number of topics was not small, the proposed method achieved better performance than LDA.
Table~\ref{tab:time} shows the training time as achieved by computers with a 2.60GHz CPU.
Since the proposed method needs to be trained by randomly sampling support and query data, its training time was long,
but not so different from that of LDAall.
The computation time of the proposed method in the test phase was 0.21, 0.03, and 0.08 seconds for the 20News, Digg, and NeurIPS data sets, respectively.
The test time was short since the prior was obtained by feeding to the trained neural networks, and the EM steps are needed for only a few documents.
Table~\ref{tab:topic} shows the estimated topics in the 20News data set.
Related vocabulary terms that were specific to the corpus were appropriately clustered.

\section{Conclusion}
\label{sec:conclusion}

We proposed a few-shot learning method that
can learn a topic model from just a small number of documents.
The proposed method uses neural networks to output Dirichlet prior parameters for topic models.
The neural networks are trained such that the topic model performs well when its parameters are estimated
by maximizing the posterior probability based on the EM algorithm.
We empirically demonstrated that the proposed method achieves better perplexity than existing topic model parameter estimation methods, and meta-learning methods. In future work,
we will use our few-shot learning approach with the EM layers and
prior distribution generating neural networks for other models,
such as Gaussian mixtures, and non-negative matrix factorization.

\bibliographystyle{abbrv}
\bibliography{aaai2021}

\end{document}